\documentclass{article}

\usepackage{color}
\usepackage{xcolor} %
\definecolor{metablue}{HTML}{0064E0}

\usepackage{spconf}
\usepackage{amsmath}
\usepackage{amssymb}
\usepackage{graphicx}
\usepackage[colorlinks=true,linkcolor=metablue,citecolor=metablue,urlcolor=metablue]{hyperref}
\usepackage{dblfloatfix} %
\usepackage{setspace} %

\usepackage{pifont} %
\usepackage{booktabs} %
\usepackage{tabularx} %
\usepackage{colortbl} %

\newcolumntype{C}[1]{>{\centering\arraybackslash}p{#1}}

\renewcommand{\paragraph}[1]{\vspace{0.5em}\noindent\textbf{#1.}~}

\newcommand{\aux}[1]{\hspace{-0.2em}\makebox[1.8em][l]{\textcolor{gray}{\tiny $\pm$\scriptsize #1}}}

\newcommand{\syncYes}{\textcolor{teal!60!black}{\ding{51}}} %
\newcommand{\syncPartial}[1]{
    \textcolor{orange!80!black}{\ding{61}}
} %
\newcommand{\syncNo}{\textcolor{red!80!black}{\ding{55}}} %

\title{Geometric Image Synchronization with Deep Watermarking}

\name{\shortstack{Pierre Fernandez$^{1,*}$, Tom\'{a}\v{s} Sou\v{c}ek$^{1,*}$, Nikola Jovanovi\'{c}$^{1,2,*,\dagger}$, 
\\ Hady Elsahar$^{1}$, Sylvestre-Alvise Rebuffi$^{1}$, Valeriu Lacatusu$^{1}$, Tuan Tran$^{1}$, Alexandre Mourachko$^{1}$}\thanks{
    \hspace{-1em}$^\star$Equal contribution. 
    $^\dagger$Contributed during an internship at FAIR.
    Correspondence to \href{mailto:pfz@meta.com}{pfz@meta.com}
}}
\address{
    $^1$FAIR, Meta Superintelligence Labs \qquad $^2$ETH Zurich
}

\usepackage{caption}
\def\1{\mathbbm{1}}

\newcommand{\etal}{et al.}

\newcommand{\ie}{i.e.,\@ }

\newcommand{\ours}{SyncSeal}

\begin{document}
\ninept

\maketitle

\begin{abstract}
Synchronization is the task of estimating and inverting geometric transformations (e.g., crop, rotation) applied to an image.
This work introduces \ours, a bespoke watermarking method for robust image synchronization, which can be applied on top of existing watermarking methods to enhance their robustness against geometric transformations. 
It relies on an embedder network that imperceptibly alters images and an extractor network that predicts the geometric transformation to which the image was subjected.
Both networks are end-to-end trained to minimize the error between the predicted and ground-truth parameters of the transformation, combined with a discriminator to maintain high perceptual quality. 
We experimentally validate our method on a wide variety of geometric and valuemetric transformations, demonstrating its effectiveness in accurately synchronizing images.
We further show that our synchronization can effectively upgrade existing watermarking methods to withstand geometric transformations to which they were previously vulnerable.\footnote{
    Code at \href{https://github.com/facebookresearch/wmar/tree/main/syncseal}{github.com/facebookresearch/wmar/tree/main/syncseal}.
}
\end{abstract}


\section{Introduction}
\label{sec:intro}

Digital watermarking is a technique used to embed information into images in a way that is imperceptible to human observers while being extractable by specialized algorithms.
An important challenge of watermarking is ensuring that embedded information remains extractable even after the host medium undergoes various transformations. 
In the image domain, geometric distortions, such as horizontal flips, rotations, cropping, and perspective changes, are particularly hard to handle, as they significantly alter the spatial arrangement of the pixels, while being common in the wild.

Robustness to geometric transformations has been addressed in prior work through various strategies.
One of those strategies, \emph{synchronization}, has historically been an important topic of watermarking research~\cite{cox2008digital}, and entails aligning the watermarked image to the coordinate system of the original image  before extraction.
This way, the original watermark itself does not need to be highly robust to geometric transformations, as the synchronization step can reliably revert them.
Traditional synchronization approaches include feature matching and homography estimation~\cite{zitova2003image, hartley2003multiple} (when the original image is known, \ie in a \emph{non-blind setting}), or embedding of a dedicated synchronization pattern~\cite{pereira2000robust, csurka1999bayesian, tirkel1998image} (when the original image is unknown, \ie in a \emph{blind setting}).
Other strategies to achieve robustness to geometric transformations rely on exhaustive search over transformation parameters, autocorrelation-based methods that detect periodic patterns, implicit synchronization, or invariant watermarking which embed the watermark in a feature space that is invariant to such transformations~\cite{cox2008digital}.
Most deep learning-based methods for watermarking rely on the invariance approach, i.e., increase geometric robustness through data augmentation during training~\cite{zhu2018hidden}. 

Despite significant research on deep learning for end-to-end watermarking, many modern methods are still vulnerable to basic geometric transformations. 
For instance, Tree-Ring~\cite{wen2023tree} can be compromised by simple translations~\cite{shamshad2025first}, and Stable Signature~\cite{fernandez2023stable} by horizontal flips. 
More recently, WMAR~\cite{jovanovic2025wmar} introduced standalone synchronization layers to make watermarking for autoregressive image generation models robust to geometric transformations, but covered only a limited family of transformations.

To improve on this, our work introduces \ours{}, a deep watermarking approach for geometric image synchronization.
\ours{} consists of an embedder network that embeds an imperceptible synchronization watermark into an image, and an extractor network that is used to estimate geometric transformation parameters of the potentially augmented watermarked image.
Both networks are trained jointly using a synchronization loss between the predicted and ground-truth transformation, combined with a discriminator to maintain high perceptual quality.

We validate our approach on a wide range of geometric and valuemetric transformations, demonstrating that \ours{} can robustly predict transformations with high accuracy while maintaining image quality.
As our main application, we further combine our synchronization model with existing watermarking methods: the synchronization watermark is embedded after the primary watermark, and we predict and invert the geometric transformation before extracting the primary watermark.
We demonstrate significant improvements in robustness against geometric transformations the methods were previously vulnerable to.

\begin{figure*}[t!]
\centering
\includegraphics[width=0.99\textwidth, clip, trim={0 7.7in 0.4in 0}]{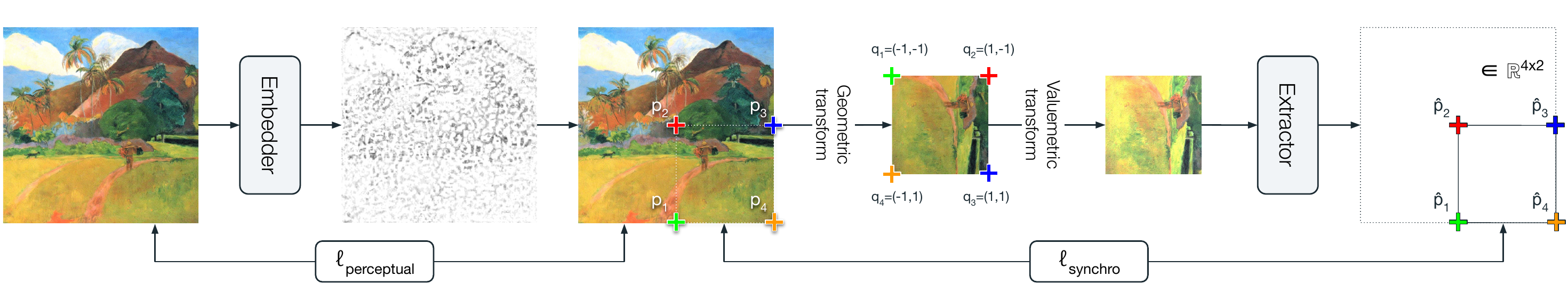}
\vspace{-0.5em}
\caption{
    Overview of the \ours\ method. The embedder generates a synchronization watermark that is imperceptibly embedded into the host image, which may undergo geometric augmentations represented as four coordinates $p_1, p_2, p_3, p_4$.
    The extractor predicts where the transformed image corners are located in the original image coordinate system, allowing for the estimation and inversion of the applied homography.
    The training optimizes the embedder and extractor using a synchronization loss and a discriminator to ensure imperceptibility.
}
\vspace{-0.5em}
\label{fig:method}
\end{figure*}

\section{Related work}
\label{sec:related}

\paragraph{Non-blind image synchronization}
In non-blind setups, where both the original and transformed images are available, classical computer vision techniques such as feature extraction with SIFT~\cite{lowe2004distinctive} and matching with RANSAC~\cite{fischler1981random}, as well as methods described in Hartley and Zisserman's book~\cite{hartley2003multiple}, are widely used for image registration and synchronization. 
In contrast, blind setups are more challenging because only the transformed image is available. This case was studied in early watermarking research~\cite[Sec. 9.3.3]{cox2008digital}.

\paragraph{Robust watermarking and synchronization}
Modern watermarking approaches typically try to achieve geometric robustness through data augmentation during training~\cite{ahmadi2020redmark, luo2020distortion, fernandez2023stable, bui2023trustmark, zhang2024editguard, sander2024watermark, fernandez2024video},
as pioneered by HiDDeN~\cite{zhu2018hidden}. 
However, some methods rely on inadequate training augmentations, such as MBRS~\cite{jia2021mbrs} and CIN~\cite{ma2022towards}, which use ``fake cropping'' where crops are pasted on black images at the same location, keeping pixel positions intact without true geometric transformation.
Some generation-time methods like Gaussian Shading~\cite{yang2024gaussian} face similar limitations, making them brittle to geometric transformations in practice.
Alternatively, exhaustive search methods systematically test transformation parameter ranges but suffer from increased computational expense and false positive rates.

\newcommand{\wangnote}{Detecting crop size but not location, which is sufficient for their tiled-block primary watermarking scheme, but not as a standalone synchronizer.}
\looseness=-1
Other approaches employ implicit synchronization through jointly-trained modules. 
DWSF~\cite{guo2023practical} propose dispersed embedding schemes with dedicated synchronization modules for block localization, 
Wang~\etal~\cite{wang2024robust} introduce a template-based method that embeds checkerboard patterns alongside watermarks and use template-enhanced networks for synchronization\footnote{\wangnote}, 
and LECA~\cite{luo2022leca} introduces a ``SyncNet'' that predicts templates to estimate scale and offset parameters. 
More recently, WMAR~\cite{jovanovic2025wmar} uses a synchronization layer as part of a watermark for autoregressive image generation, recovering image tokens at their correct positions.
It employs localized watermarking~\cite{sander2024watermark} to embed four binary messages in image corners and find separating lines to recover rotation, horizontal flips and crops. 
However, all these approaches have significant limitations.
They are not robust against the full range of geometric transformations (see Tab.~\ref{tab:sync_capabilities}) and/or depend on template-matching algorithms.
This results in reduced efficiency and accuracy, particularly for complex transformations. 
For instance, perspective changes involve many degrees of freedom and become computationally prohibitive to search exhaustively.
Our approach directly predicts the geometric mapping, resulting in improved generalizability and faster computation.

\begin{table}[t!]
    \centering
    \caption{
        Comparison of synchronization capabilities across methods: 
        full (\syncYes), partial (\syncPartial{}), or not at all (\syncNo).
        Standalone indicates that the synchronization module can be used as a separate module.
    }
    \vspace{-0.5em}
    \renewcommand{\arraystretch}{0.92}
    \resizebox{\columnwidth}{!}{
        \setlength{\tabcolsep}{3pt}
        \begin{tabular}{l *{5}{C{0.15\columnwidth}}}
            \toprule
            Method                      & Standalone & Rotation & H. Flip & Crop & Perspective \\
            \midrule
            LECA~\cite{luo2022leca}             & \syncYes & \syncNo & \syncNo & \syncYes & \syncNo\\
            DWSF~\cite{guo2023practical}        & \syncNo & \syncPartial{($\pm30^\circ$)} & \syncNo & \syncYes & \syncNo \\
            Wang~\etal~\cite{wang2024robust}\footnotemark[\value{footnote}]    & \syncPartial{} & \syncYes & \syncNo  & \syncPartial{} & \syncNo \\
            WMAR~\cite{jovanovic2025wmar}       & \syncYes & \syncPartial{($\pm20^\circ$ and $90^\circ$)} & \syncYes & \syncPartial{($>0.5$ in area)} & \syncNo \\
            \ours{} (ours)                      & \syncYes & \syncYes    & \syncYes & \syncYes & \syncYes \\
            \bottomrule
        \end{tabular}
    }
\vspace{-1em}
\label{tab:sync_capabilities}
\end{table}

\section{Method}
\label{sec:method}

In this section, we present the \ours\ framework for image synchronization. 
We first describe how geometric transformations are represented as homographies and corner correspondences (Sec.~\ref{sec:geometric_representation}). 
We then detail the inference pipeline (Sec.~\ref{sec:pipeline}), followed by the end-to-end training (Sec.~\ref{sec:training}) and the implementation details (Sec.~\ref{sec:implem}).

\subsection{Representation of the geometric transformations}
\label{sec:geometric_representation}

Common image geometric transformations can be represented by homography matrices $\mathbf{H}\in\mathbb{R}^{3 \times 3}$ with 8 degrees of freedom ($\mathbf{H}_{33}=1$).
The homography matrix maps a (normalized) pixel location in the original image $p\in[-1,1]^2$ to its corresponding location in the augmented image $q\in\mathbb{R}^2$, potentially outside of the visible $[-1,1]^2$ region, as follows:
\begin{equation}
\begin{bmatrix} q_x \\ q_y \\ 1 \end{bmatrix} \sim \mathbf{H} \begin{bmatrix} p_x \\ p_y \\ 1 \end{bmatrix},
\end{equation}
where $\sim$ denotes equality up to a nonzero scale factor.
The matrix is uniquely determined by four non-collinear corresponding point pairs. Therefore, instead of predicting the homography matrix, in this work, we choose to directly predict the four points $\{p_i\}_{i=1}^4$ in the original image corresponding to the corner points of the transformed image $(q_1, q_2, q_3, q_4)$$=$$((-1,-1), (1,-1), (1,1), (-1,1))$.
By recovering these four correspondences, we can fully reconstruct the applied geometric transformation and invert it to restore the original image coordinate system.
For instance, as shown in \autoref{fig:method}, a crop of the bottom-right corner followed by a 90$^\circ$ rotation would be represented by $(p_1, p_2, p_3, p_4)$$=$$((0,1), (0,0), (1,0), (1,1))$.

\subsection{Inference pipeline}
\label{sec:pipeline}

The \emph{embedder} imperceptibly embeds a synchronization watermark into the host image, which may then be subjected to geometric transformations. 
The \emph{extractor} predicts where the transformed image corners are located in the original image coordinate system, which allows for the estimation and inversion of the applied homography.

\paragraph{Embedder}\label{par:embedder} 
The embedder $\text{Emb}$ is responsible for imperceptibly embedding a synchronization watermark into the host image. 
It takes an image $\mathbf{I} 
\in [0,1]^{3 \times H \times W}$ as input and outputs a synchronization watermark $\mathbf{w} 
\in [-1, 1]^{3 \times H \times W}$, which is then summed with the input image to produce $\mathbf{I}_w$.
To ensure imperceptibility, we constrain $\text{Emb}$ with a tanh activation function and only embed on the luminance channel of the image. 
We also use a Just Noticeable Difference (JND) map which attenuates the watermark based on per-pixel sensitivity to changes, similar to prior works \cite{zhang2008just, sander2024watermark}.
The overall embedding process can therefore be expressed as:
\begin{equation}\label{eq:watermark_image}
\mathbf{I}_w = \mathbf{I} 
+ \alpha_w \cdot \mathbf{w}, \quad \mathbf{w} 
= \text{JND}( \mathbf{I}) \odot \text{tanh} \left( \text{Emb}(\mathbf{I}) \right),
\end{equation}
where $\odot$ denotes element-wise multiplication and
$\alpha_w \in \mathbb{R}^+$ is a scaling factor that controls the strength of the watermark. 

\paragraph{Extractor}\label{par:extractor} 
The extractor $\text{Ext}$ is tasked with recovering the geometric synchronization information from the potentially transformed image~$\tilde{\mathbf{I}}_w$. 
The extractor takes $\tilde{\mathbf{I}}_w$ and outputs the predicted 4-point coordinates $\hat{p}_1, \hat{p}_2, \hat{p}_3, \hat{p}_4$ (as a vector of $\mathbb{R}^8$) that represent where the transformed image corners $q_1, q_2, q_3, q_4$ are located in the original image coordinate system. 
These predicted correspondences allow us to obtain the homography parameters for synchronization (see Sec.~\ref{sec:geometric_representation}), and to invert the transformation.

\paragraph{Operating at different resolutions}
When processing an arbitrary resolution image, it is first downsampled to a fixed size $H \times W$ using bilinear interpolation and processed by the embedder to produce the synchronization watermark $\mathbf{w}$. 
Then, $\mathbf{w}$ is upsampled back to $H_{ori} 
\times W_{ori}$ and summed with the original image to obtain the watermarked $\mathbf{I}_w$.
Similarly, $\tilde{\mathbf{I}}_w$ is resized to $H \times W$ before extraction.

Note that our method operates at any resolution but cannot predict scale changes (resize transformations). 
This means it cannot determine the original resolution of an image that has been resized.
This is not an issue for modern watermarking methods that operate at fixed resolutions~\cite{wen2023tree, yang2024gaussian, jovanovic2025wmar, bui2023trustmark,zhang2024editguard,sander2024watermark,fernandez2024video}, but would be needed for watermarks that are not resolution-agnostic such as the original HiDDeN~\cite{zhu2018hidden}, or Tree-Ring and WMAR used on generative models at different resolutions.

\subsection{Training}
\label{sec:training}

We end-to-end optimize the embedder and the extractor to learn robust synchronization under geometric and valuemetric distortions. 
During training, an original image $\mathbf{I}$ is first passed through the embedder to produce the watermarked image $\mathbf{I}_w$. 
$\mathbf{I}_w$ then undergoes geometric augmentations, producing both the augmented image and the ground-truth corner locations, followed by valuemetric transformations, to simulate real-world image editing, resulting in an augmented image $\tilde{\mathbf{I}}_{w}$.
Finally, $\tilde{\mathbf{I}}_{w}$ is fed into the extractor which predicts the geometric transformation represented by the 4-point coordinates $\hat{p}_1, \hat{p}_2, \hat{p}_3, \hat{p}_4$.
We ensure accurate synchronization and high imperceptibility by combining a synchronization loss with a discriminator.

\vspace{0.5em}\noindent
The \textbf{geometric augmenter} applies a sequence of geometric transformations to the watermarked image $\mathbf{I}_w$ including identity, rotation, cropping, perspective distortion, and horizontal flips. 
For each applied transformation, the geometric augmenter computes where the transformed image corners $q_1, q_2, q_3, q_4$ belong in the original image coordinate system, yielding ground-truth points $p_1, p_2, p_3, p_4$ that are used for supervision of the synchronization task.

\vspace{0.5em}\noindent
The \textbf{valuemetric augmenter} applies additional distortions to the image, such as brightness, contrast, JPEG compression, etc.
This makes the synchronization robust to common image manipulations.

\vspace{0.5em}\noindent
The \textbf{loss function} $
\mathcal{L} = 
\mathcal{L}_{\text{sync}} 
+ \lambda_{\text{adv}} \mathcal{L}_{\text{adv}}
$ comprises a synchronization loss and an adversarial loss with a weighting factor $\lambda_{\text{adv}}$.
The synchronization loss $\mathcal{L}_{\text{sync}} = \sum_{i=1}^{4} \|\hat{p}_i - p_i\|_1$ is computed as the $\ell_1$ distance between the extractor's predicted 4-point coordinates $\hat{p}_1, \hat{p}_2, \hat{p}_3, \hat{p}_4$ and the ground-truth coordinates $p_1, p_2, p_3, p_4$ provided by the geometric augmenter.
This loss term explicitly drives the extractor to accurately predict where the transformed image corners are located in the original image coordinate system (we use $\ell_1$ instead of $\ell_2$ as it favors sparsity and gave better results in practice).
To improve the imperceptibility of the embedded watermark, we employ an adversarial loss $\mathcal{L}_{\text{adv}} = -D(\mathbf{I}_w)$ using a patch-based discriminator $D$. It maximizes the likelihood that a watermarked image is labeled as non-watermarked by $D$. 
In a distinct optimization step, $D$ is trained to distinguish between real and watermarked images by minimizing a hinge loss.
All details are available in~\cite{rombach2022high}, whose discriminator we directly adopt.

\subsection{Implementation details}\label{sec:implem}

For the embedder, we utilize a U-Net based architecture~\cite{saharia2022photorealistic} with 3.3M parameters. 
The extractor is ConvNext-v2 Tiny~\cite{woo2023convnext} with 27.9M parameters.
The models are trained on SA-1b~\cite{kirillov2023segment} with images resized to 256$\times$256 pixels. 
Optimization is performed using AdamW with learning rate $10^{-3}$. 
We train for 300.000 iterations with batch size 192.
We set $\alpha_w = 0.2$ and $\lambda_{\text{adv}} = 0.1$.
This takes approximately two days on 8 V100 16Gb GPUs.

Each image randomly undergoes 3 geometric transformations (among identity, crop, horizontal flip, rotation, perspective) and 2 valuemetric transformations (JPEG, Gaussian blur, brightness, contrast, saturation). 
Crop area ranges from 0.3 to 1.0 of the original; rotation is between $\pm135^\circ$; maximum perspective scale is 0.3 (relative displacement of a corner point along each edge). 
JPEG quality varies from 40-80; brightness, contrast, saturation from 0.5-2.0; and blur from kernel size 1 to 9 (with torchvision default parameters).

\section{Experiments}
\label{sec:experiments}

\subsection{Image quality}
\label{sec:quality}

\begin{figure}[b!]
\centering
\footnotesize
\begin{tabular*}{\columnwidth}{@{\extracolsep{\fill}}c@{\hspace{0.02\columnwidth}}c@{\hspace{0.02\columnwidth}}c}
\includegraphics[width=0.32\columnwidth]{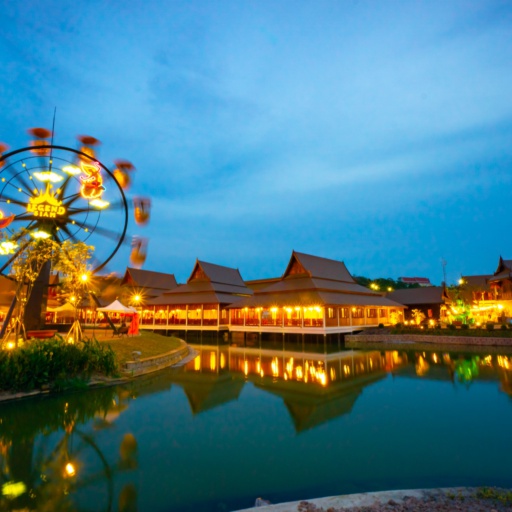} &
\includegraphics[width=0.32\columnwidth]{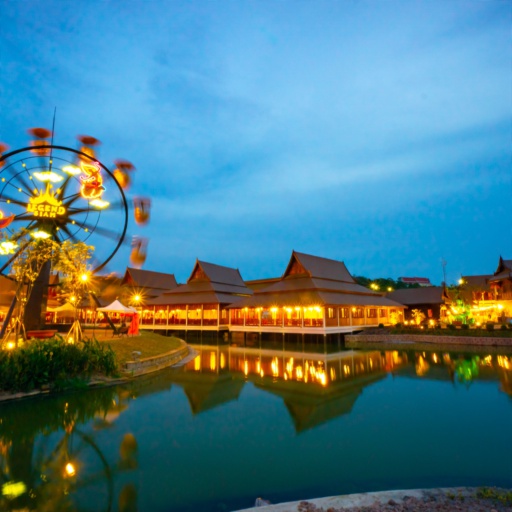} &
\includegraphics[width=0.32\columnwidth]{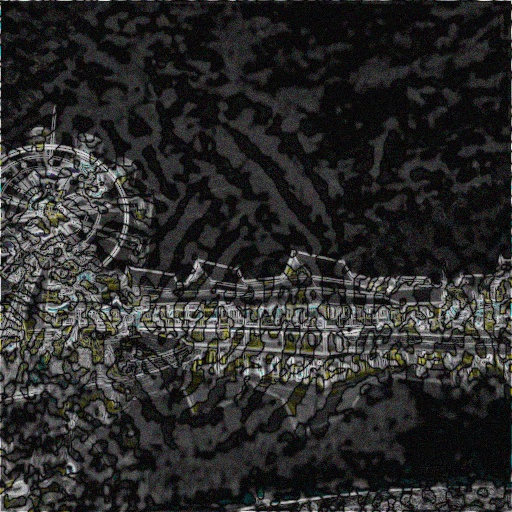} \\
Original & Watermarked & Difference ($\times$10)
\end{tabular*}
\vspace{-0.5em}
\caption{
Visual comparison of original and watermarked images (at 512$\times$512). 
The difference (amplified $\times$10) is barely visible and mainly in textured areas, which makes the watermark imperceptible.
}
\label{fig:quality_comparison}
\end{figure}

We evaluate the visual quality of images watermarked with \ours\ on a test set with 1000 images held-out from SA-1b resized to the resolution of 512$\times$512 pixels.
Specifically, we measure PSNR and SSIM to assess both pixel-level fidelity and perceptual quality.
We observe high metrics, and hard-to-perceive watermarking artifacts, which are mainly located in textured areas of the image due to the JND and discriminator mechanisms.
Quantitative results are reported in \autoref{tab:image_quality}, while \autoref{fig:quality_comparison} shows a qualitative example.

\subsection{Synchronization accuracy and robustness}
\label{sec:robustness}

\begin{figure}[b!]
\centering
\includegraphics[width=\columnwidth]{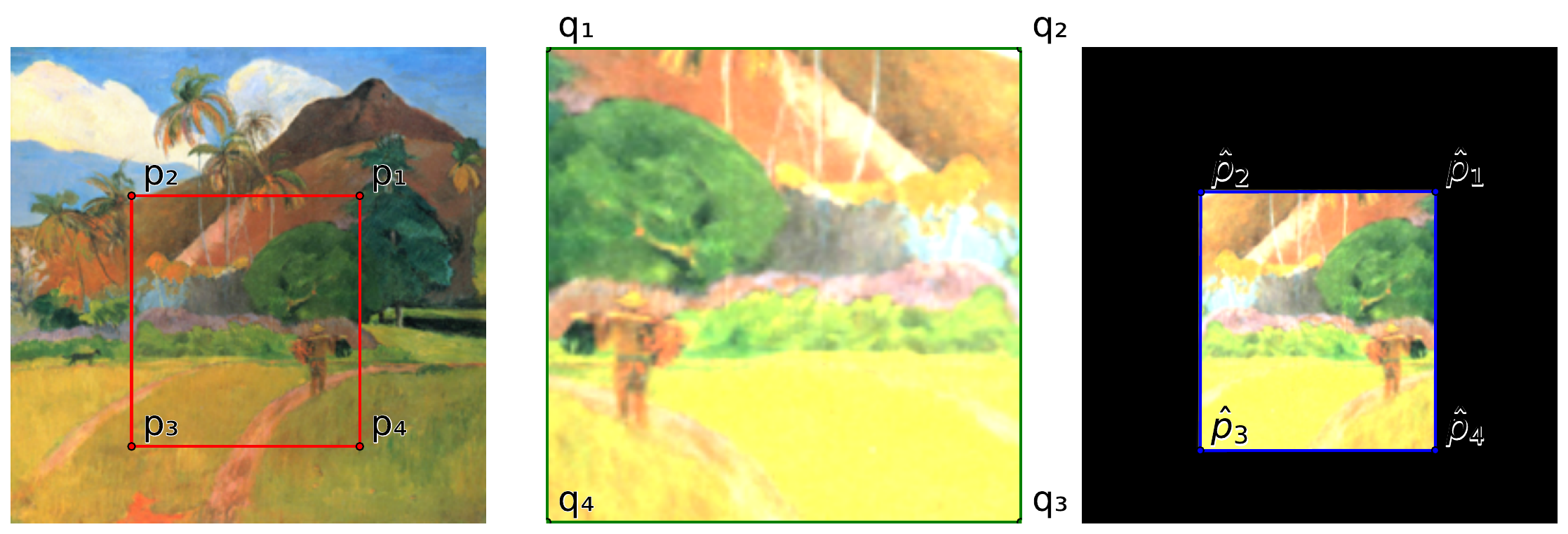}
\footnotesize
\begin{tabularx}{\columnwidth}{XXX}
    \centering Original & \centering Augmented & \centering Resynchronized
\end{tabularx}
\vspace{-1.5em}
\caption{
    Example of corner prediction on a transformed image. 
    The close alignment between ground truth and predicted corners demonstrates accurate synchronization under combined geometric and valuemetric transformations.
}
\label{fig:augmentations}
\end{figure}

We evaluate the synchronization accuracy of our method on images subjected to various geometric and valuemetric transformations. 
\autoref{fig:augmentations} shows a qualitative example demonstrating our extractor's ability to accurately predict corner locations $p_1, p_2, p_3, p_4$ for an image that has undergone geometric and valuemetric transformations.
Quantitatively, we measure the average $\ell_2$ distance between the predicted corner locations $\hat{p}_i$ and the ground-truth locations $p_i$ on our test set.
In \autoref{tab:synchronization_accuracy}, we see that most of the time the average error is below 10 pixel distance, with some exceptions for harder augmentations such as low contrast and perspective distortions.

\begin{table}[t!]
\centering
\caption{
    Average corner distance error (in pixels, lower is better) of our method under geometric and valuemetric augmentations on the test set.
}\vspace{-0.4em}
\resizebox{\columnwidth}{!}{
\setlength{\tabcolsep}{1.5pt}
\renewcommand{\arraystretch}{1.2}
\begin{tabular}{l@{~~~} *{8}{r}}
\toprule
& \multicolumn{7}{c}{Geomometric Augmentations} & \\
Val. Aug. & \multicolumn{1}{c}{Iden.} & \multicolumn{1}{c}{HFlip} & \multicolumn{1}{c}{Rot 5} & \multicolumn{1}{c}{Rot 90} & \multicolumn{1}{c}{Crop 0.5} & \multicolumn{1}{c}{Crop 0.9} & \multicolumn{1}{c}{Persp 0.3} & Average \\
\cmidrule(lr){2-9}
Ident. & 0.5~\aux{1.6} & 3.3~\aux{2.1} & 1.3~\aux{0.2} & 3.9~\aux{3.1} & 7.2~\aux{4.4} & 3.9~\aux{2.1} & 8.3~\aux{10.6} & 4.1~\aux{5.4} \\
Gray & 0.4~\aux{1.5} & 3.2~\aux{2.2} & 1.3~\aux{0.2} & 3.9~\aux{3.1} & 7.1~\aux{4.4} & 3.9~\aux{2.1} & 8.3~\aux{10.7} & 4.0~\aux{5.4} \\
Hue 0.1 & 0.8~\aux{2.1} & 3.6~\aux{2.3} & 1.3~\aux{0.2} & 4.5~\aux{3.5} & 7.7~\aux{4.8} & 4.4~\aux{2.5} & 8.8~\aux{10.6} & 4.4~\aux{5.6} \\
Bright. 0.5 & 0.4~\aux{1.5} & 2.5~\aux{2.4} & 1.3~\aux{0.4} & 3.2~\aux{3.0} & 7.4~\aux{4.6} & 4.4~\aux{2.6} & 8.5~\aux{10.1} & 4.0~\aux{5.4} \\
Bright. 2.0 & 2.8~\aux{12.3} & 6.0~\aux{6.3} & 1.3~\aux{0.2} & 9.3~\aux{30.3} & 14.5~\aux{19.6} & 6.8~\aux{7.2} & 13.9~\aux{24.9} & 7.8~\aux{18.2} \\
Contr. 0.5 & 0.4~\aux{1.5} & 3.9~\aux{2.1} & 27.6~\aux{4.4} & 7.2~\aux{4.3} & 7.5~\aux{4.6} & 4.4~\aux{2.7} & 90.2~\aux{50.6} & 20.2~\aux{35.5} \\
Contr. 2.0 & 1.4~\aux{3.7} & 4.4~\aux{3.9} & 1.6~\aux{1.5} & 8.7~\aux{19.6} & 11.7~\aux{10.0} & 5.8~\aux{3.8} & 14.4~\aux{18.8} & 6.9~\aux{12.1} \\
JPEG 40 & 0.3~\aux{1.4} & 2.8~\aux{2.6} & 1.5~\aux{1.2} & 4.0~\aux{3.8} & 8.6~\aux{6.2} & 5.3~\aux{3.7} & 13.8~\aux{19.2} & 5.2~\aux{9.1} \\
G.Blur 9 & 0.0~\aux{0.1} & 4.7~\aux{2.3} & 7.1~\aux{4.4} & 11.5~\aux{14.4} & 8.5~\aux{6.4} & 5.1~\aux{4.0} & 28.8~\aux{32.2} & 9.4~\aux{16.2} \\
\midrule
Average & 0.8~\aux{4.6} & 3.8~\aux{3.3} & 4.9~\aux{8.5} & 6.3~\aux{13.6} & 8.9~\aux{8.9} & 4.9~\aux{3.8} & 21.7~\aux{35.0} & 7.3~\aux{16.4} \\
\bottomrule
\end{tabular}
}
\label{tab:synchronization_accuracy}
\end{table}

\subsection{Comparison to other synchronization methods}
\label{sec:comparison}
We compare our method against 
a classical registration approach using RANSAC-based SIFT feature matching~\cite{lowe2004distinctive} %
as implemented in OpenCV~\cite{opencv_library} (non-blind, requiring the original image) 
and the WMAR synchronization~\cite{jovanovic2025wmar} (blind), which uses WAM~\cite{sander2024watermark} to embed binary messages in image quadrants and employs search algorithms to recover rotation, horizontal flips, and crops. 
Methods are evaluated under identical geometric and valuemetric augmentations except for WMAR, for which we only evaluate top-left crops, since the method does not support other crop types. 
The average corner distance is reported in \autoref{tab:method_comparison}.
Our method consistently outperforms the blind baseline across all geometric transformations.
WMAR shows limited capability, being restricted to specific transformation types and exhibiting higher error rates due to its template-matching-based approach.
Our direct prediction approach is also faster to compute (0.01s) compared to WMAR (0.6s) using a vanilla PyTorch implementation on a V100 GPU. 
Finally, somewhat surprisingly, our method remains competitive with SIFT, a method that requires the original reference image (that is not available in a common watermarking application).

\begin{table}[t!]
\centering
\caption{
    Comparison with other synchronization methods on the test set.%
}\vspace{-0.4em}
\label{tab:image_quality}
\label{tab:method_comparison}
\resizebox{\columnwidth}{!}{
\setlength{\tabcolsep}{2.5pt}
\renewcommand{\arraystretch}{1.2}
\begin{tabular}{l c c c c r r r r r}
\toprule
& \multicolumn{2}{c}{Quality ($\uparrow$)} & \multicolumn{2}{c}{Time (ms) ($\downarrow$)} & \multicolumn{4}{c}{Average corner distance ($\downarrow$)} \\
\cmidrule(lr){2-3}\cmidrule(lr){4-5}\cmidrule(lr){6-9}
Method & PSNR & SSIM & Embed & Extract & Hflip & Rotate 10 & Crop 0.7 & Persp. 0.1 \\
\midrule
SIFT & -- & -- & -- & 146 & 1.5 & 1.8 & 1.6 & 1.5 \\
WMAR & 35.2 & 0.971 & 69 & 565 & 1.7 & 5.5 & 55.1 & 49.3 \\
\rowcolor{teal!8} \ours & 43.9 & 0.994 & 12 & 12 & 3.3 & 0.5 & 3.8 & 6.6 \\
\bottomrule
\end{tabular}
}
\end{table}

\subsection{Applications to existing watermarking methods}
\label{sec:applications}

We combine \ours\ with existing watermarking methods to enhance their robustness against geometric transformations. 
The synchronization watermark is embedded after the primary watermark, and geometric transformation is predicted and inverted before primary watermark extraction.

For post-hoc watermarking methods, 
we evaluate \ours{} with 
MBRS~\cite{jia2021mbrs} and 
CIN~\cite{ma2022towards} 
which suffer from poor geometric robustness due to inadequate training augmentations, and the more recent methods TrustMark~\cite{bui2023trustmark} and WAM~\cite{sander2024watermark}, which are more robust but still vulnerable to strong transformations.
We evaluate the bit accuracy of the extracted message under geometric transformations.
For generation-time watermarking methods, 
we evaluate \ours{} as a replacement for WMAR's synchronization module~\cite{jovanovic2025wmar} and with Tree-Ring~\cite{wen2023tree}, which is vulnerable to geometric transformations because it embeds the watermark in the initial noise of the diffusion process.
For WMAR and for Tree-Ring, we generate $1000$ images conditioned on ImageNet classes as in~\cite{jovanovic2025wmar}, with resolution $256 \times 256$ for both.
We report True Positive Rate (TPR) at a fixed False Positive Rate (FPR), as these methods focus on watermark detection rather than message extraction. 

Results in~\autoref{tab:encoderdecoder_results} show substantial improvements for both kinds of methods: TrustMark and WAM achieve over $97\%$ bit accuracy with our synchronization, and Tree-Ring shows consistent detection improvements across all geometric transformations.
While the original synchronization method of WMAR supports horizontal flips, rotations of up to $20$ degrees, and top-left crops, replacing it with \ours{} greatly broadens the range of supported transformations.
For example, its TPR under $30^\circ$ rotation increases from $1\%$ to $99.8\%$, and under perspective shifts with parameter $0.3$ from $1\%$ to $95\%$.

\begin{table}[t!]
\centering
\caption{
    Results for various watermarking methods with (\syncYes) and without (\syncNo) our robust SyncSeal synchronization. SyncSeal significantly improves the robustness of both post-hoc and generation-time watermarking methods.
}\vspace{-0.4em}
\label{tab:encoderdecoder_results}
\resizebox{\columnwidth}{!}{
\begin{tabular}{l c *{6}{p{0.12\columnwidth}}}
\toprule
Method & Sync & Identity & Hflip & Rot.~30 & Crop~0.5 & Persp.~0.3 & Average \\
\midrule
\multicolumn{8}{l}{ \hspace{-4pt}\textcolor{black}{\textit{\footnotesize Encoder-decoder methods (Bit accuracy) - 512$\times$512 images from SA-1b}}} \\
CIN & \syncNo & 1.00 & 0.51 & 0.49 & 0.50 & 0.50 & 0.60 \\
\rowcolor{teal!8} CIN & \syncYes & 1.00 & 0.98 & 1.00 & 0.83 & 0.84 & 0.93 \\
[0.2em]MBRS & \syncNo & 0.98 & 0.50 & 0.50 & 0.50 & 0.50 & 0.60 \\
\rowcolor{teal!8} MBRS & \syncYes & 0.95 & 0.75 & 0.88 & 0.57 & 0.63 & 0.76 \\
[0.2em]TrustMark & \syncNo & 1.00 & 1.00 & 0.50 & 0.53 & 0.51 & 0.71 \\
\rowcolor{teal!8} TrustMark & \syncYes & 0.99 & 0.99 & 0.99 & 0.92 & 0.97 & 0.97 \\
[0.2em]WAM & \syncNo & 1.00 & 0.99 & 0.50 & 0.99 & 0.68 & 0.83 \\
\rowcolor{teal!8} WAM & \syncYes & 1.00 & 1.00 & 1.00 & 0.97 & 0.98 & 0.99 \\
\midrule
\multicolumn{8}{l}{ \hspace{-4pt}\textcolor{black}{\textit{\footnotesize Generation-time methods (TPR@FPR=1\%) - 256$\times$256 AI-generated images}}} \\
WMAR &  \textcolor{red!80!black}{{\scriptsize{WAM}}} & 1.00 & 1.00 & 0.01 & 0.79 & 0.01 & 0.56 \\
\rowcolor{teal!8} WMAR & \syncYes & 1.00 & 1.00 & 1.00 & 0.92 & 0.95 & 0.97 \\
[0.2em]Tree-Ring & \syncNo & 1.00 & 0.90 & 0.00 & 0.01 & 0.02 & 0.39 \\
\rowcolor{teal!8} Tree-Ring & \syncYes & 0.99 & 0.95 & 0.94 & 0.72 & 0.77 & 0.88 \\
\bottomrule
\end{tabular}
}
\end{table}

\section{Conclusion}
\label{sec:conclusion}
We presented an active geometric synchronization framework that end-to-end learns to predict the parameters of geometric transformations applied to an image.
Our experiments demonstrate strong synchronization accuracy and show how it can enhance geometric robustness of other watermarking methods.

Despite these advantages, our approach has limitations consistent with traditional synchronization methods~\cite{cox2008digital}: 
inaccuracies can arise both from the synchronization module's geometric prediction and the subsequent watermark extractor, potentially compounding errors;
the approach increases computational complexity compared to direct watermark extraction, requiring separate embedding and extraction of synchronization information.
Future work could investigate the fundamental trade-offs between geometric robustness and watermark capacity when using this kind of synchronization module.

\clearpage

\bibliographystyle{IEEEbib}
\bibliography{references}

\begin{thebibliography}{10}

\bibitem{cox2008digital}
Ingemar~J Cox, Matthew~L Miller, Jeffrey~A Bloom, Jessica Fridrich, and Ton
  Kalker,
\newblock ``Digital watermarking,''
\newblock {\em Morgan Kaufmann Publishers}, vol. 54, pp. 56--59, 2008.

\bibitem{zitova2003image}
Barbara Zitova and Jan Flusser,
\newblock ``Image registration methods: a survey,''
\newblock {\em Image and vision computing}, 2003.

\bibitem{hartley2003multiple}
Richard Hartley and Andrew Zisserman,
\newblock {\em Multiple view geometry in computer vision},
\newblock Cambridge university press, 2003.

\bibitem{pereira2000robust}
Shelby Pereira and Thierry Pun,
\newblock ``Robust template matching for affine resistant image watermarks,''
\newblock {\em IEEE TIP}, 2000.

\bibitem{csurka1999bayesian}
Gabriella Csurka, Fr{\'e}d{\'e}ric Deguillaume, Joseph~JK {\'O}~Ruanaidh, and
  Thierry Pun,
\newblock ``A bayesian approach to affine transformation resistant image and
  video watermarking,''
\newblock in {\em International Workshop on Information Hiding}. Springer,
  1999.

\bibitem{tirkel1998image}
Andrew~Z Tirkel, Charles~F Osborne, and Thomas~E Hall,
\newblock ``Image and watermark registration,''
\newblock {\em Signal processing}, vol. 66, no. 3, pp. 373--383, 1998.

\bibitem{zhu2018hidden}
Jiren Zhu, Russell Kaplan, Justin Johnson, and Li~Fei-Fei,
\newblock ``Hidden: Hiding data with deep networks,''
\newblock in {\em ECCV}, 2018.

\bibitem{wen2023tree}
Yuxin Wen, John Kirchenbauer, Jonas Geiping, and Tom Goldstein,
\newblock ``Tree-ring watermarks: Fingerprints for diffusion images that are
  invisible and robust,''
\newblock {\em NeurIPS}, 2023.

\bibitem{shamshad2025first}
Fahad Shamshad, Tameem Bakr, Yahia~Salaheldin Shaaban, Noor~Hazim Hussein,
  Karthik Nandakumar, and Nils Lukas,
\newblock ``First-place solution to neurips 2024 invisible watermark removal
  challenge,''
\newblock in {\em Workshop on GenAI Watermarking (ICLR)}, 2025.

\bibitem{fernandez2023stable}
Pierre Fernandez, Guillaume Couairon, Herv{\'e} J{\'e}gou, Matthijs Douze, and
  Teddy Furon,
\newblock ``The stable signature: Rooting watermarks in latent diffusion
  models,''
\newblock {\em ICCV}, 2023.

\bibitem{jovanovic2025wmar}
Nikola Jovanovi\'{c}, Ismail Labiad, Tom\'{a}\v{s} Sou\v{c}ek, Martin Vechev,
  and Pierre Fernandez,
\newblock ``Watermarking autoregressive image generation,''
\newblock {\em arXiv preprint arXiv:2506.16349}, 2025.

\bibitem{lowe2004distinctive}
David~G Lowe,
\newblock ``Distinctive image features from scale-invariant keypoints,''
\newblock {\em IJCV}, 2004.

\bibitem{fischler1981random}
Martin~A Fischler and Robert~C Bolles,
\newblock ``Random sample consensus: a paradigm for model fitting with
  applications to image analysis and automated cartography,''
\newblock {\em Communications of the ACM}, vol. 24, no. 6, pp. 381--395, 1981.

\bibitem{ahmadi2020redmark}
Mahdi Ahmadi, Alireza Norouzi, Nader Karimi, Shadrokh Samavi, and Ali Emami,
\newblock ``Redmark: Framework for residual diffusion watermarking based on
  deep networks,''
\newblock {\em Expert Systems with Applications}, 2020.

\bibitem{luo2020distortion}
Xiyang Luo, Ruohan Zhan, Huiwen Chang, Feng Yang, and Peyman Milanfar,
\newblock ``Distortion agnostic deep watermarking,''
\newblock in {\em CVPR}, 2020.

\bibitem{bui2023trustmark}
Tu~Bui, Shruti Agarwal, and John Collomosse,
\newblock ``Trustmark: Universal watermarking for arbitrary resolution
  images,''
\newblock {\em arXiv preprint arXiv:2311.18297}, 2023.

\bibitem{zhang2024editguard}
Xuanyu Zhang, Runyi Li, Jiwen Yu, Youmin Xu, Weiqi Li, and Jian Zhang,
\newblock ``Editguard: Versatile image watermarking for tamper localization and
  copyright protection,''
\newblock in {\em Proceedings of the IEEE/CVF Conference on Computer Vision and
  Pattern Recognition}, 2024, pp. 11964--11974.

\bibitem{sander2024watermark}
Tom Sander, Pierre Fernandez, Alain Durmus, Teddy Furon, and Matthijs Douze,
\newblock ``Watermark anything with localized messages,''
\newblock {\em ICLR}, 2025.

\bibitem{fernandez2024video}
Pierre Fernandez, Hady Elsahar, I~Zeki Yalniz, and Alexandre Mourachko,
\newblock ``Video seal: Open and efficient video watermarking,''
\newblock {\em arXiv preprint arXiv:2412.09492}, 2024.

\bibitem{jia2021mbrs}
Zhaoyang Jia, Han Fang, and Weiming Zhang,
\newblock ``Mbrs: Enhancing robustness of dnn-based watermarking by mini-batch
  of real and simulated jpeg compression,''
\newblock in {\em ACM Multimedia Conference}, 2021, pp. 41--49.

\bibitem{ma2022towards}
Rui Ma, Mengxi Guo, Yi~Hou, Fan Yang, Yuan Li, Huizhu Jia, and Xiaodong Xie,
\newblock ``Towards blind watermarking: Combining invertible and non-invertible
  mechanisms,''
\newblock in {\em ACM Multimedia Conference}, 2022, pp. 1532--1542.

\bibitem{yang2024gaussian}
Zijin Yang, Kai Zeng, Kejiang Chen, Han Fang, Weiming Zhang, and Nenghai Yu,
\newblock ``Gaussian shading: Provable performance-lossless image watermarking
  for diffusion models,''
\newblock in {\em CVPR}, 2024.

\bibitem{guo2023practical}
Hengchang Guo, Qilong Zhang, Junwei Luo, Feng Guo, Wenbin Zhang, Xiaodong Su,
  and Minglei Li,
\newblock ``Practical deep dispersed watermarking with synchronization and
  fusion,''
\newblock in {\em ACM Multimedia Conference}, 2023, pp. 7922--7932.

\bibitem{wang2024robust}
Ke~Wang, Shaowu Wu, Xiaolin Yin, Wei Lu, Xiangyang Luo, and Rui Yang,
\newblock ``Robust image watermarking with synchronization using template
  enhanced-extracted network,''
\newblock {\em IEEE TCSVT}, 2024.

\bibitem{luo2022leca}
Xiyang Luo, Michael Goebel, Elnaz Barshan, and Feng Yang,
\newblock ``Leca: A learned approach for efficient cover-agnostic
  watermarking,''
\newblock {\em arXiv preprint arXiv:2206.10813}, 2022.

\bibitem{zhang2008just}
Xiaohui Zhang, Weisi Lin, and Ping Xue,
\newblock ``Just-noticeable difference estimation with pixels in images,''
\newblock {\em JVCIR}, vol. 19, no. 1, pp. 30--41, 2008.

\bibitem{rombach2022high}
Robin Rombach, Andreas Blattmann, Dominik Lorenz, Patrick Esser, and Bj{\"o}rn
  Ommer,
\newblock ``High-resolution image synthesis with latent diffusion models,''
\newblock in {\em CVPR}, 2022.

\bibitem{saharia2022photorealistic}
Chitwan Saharia, William Chan, Saurabh Saxena, Lala Li, Jay Whang, Emily~L
  Denton, Kamyar Ghasemipour, Raphael Gontijo~Lopes, Burcu Karagol~Ayan, Tim
  Salimans, et~al.,
\newblock ``Photorealistic text-to-image diffusion models with deep language
  understanding,''
\newblock {\em NeurIPS}, 2022.

\bibitem{woo2023convnext}
Sanghyun Woo, Shoubhik Debnath, Ronghang Hu, Xinlei Chen, Zhuang Liu, In~So
  Kweon, and Saining Xie,
\newblock ``Convnext v2: Co-designing and scaling convnets with masked
  autoencoders,''
\newblock in {\em CVPR}, 2023.

\bibitem{kirillov2023segment}
Alexander Kirillov, Eric Mintun, Nikhila Ravi, Hanzi Mao, Chloe Rolland, Laura
  Gustafson, Tete Xiao, Spencer Whitehead, Alexander~C Berg, Wan-Yen Lo,
  et~al.,
\newblock ``Segment anything,''
\newblock in {\em ICCV}, 2023.

\bibitem{opencv_library}
G.~Bradski,
\newblock ``{The OpenCV Library},''
\newblock {\em Dr. Dobb's Journal of Software Tools}, 2000.

\end{thebibliography}

\end{document}